\documentclass{article}

\usepackage{arxiv}  
\usepackage[utf8]{inputenc} % allow utf-8 input
\usepackage[T1]{fontenc}    % use 8-bit T1 fonts
\usepackage{hyperref}       % hyperlinks
\usepackage{url}            % simple URL typesetting
\usepackage{booktabs}       % professional-quality tables
\usepackage{amsfonts}       % blackboard math symbols
\usepackage{nicefrac}       % compact symbols for 1/2, etc.
\usepackage{microtype}      % microtypography
\usepackage{lipsum}
\usepackage{graphicx}
\usepackage{amsmath}
\usepackage{multirow}   % 支持 \multirow
\usepackage{booktabs}   % 支持 \toprule 等
\usepackage{graphicx}   % 支持 \resizebox
\usepackage{makecell}  
\usepackage{graphicx}
\usepackage{tcolorbox}
\usepackage{xcolor}
\tcbuselibrary{skins}
\definecolor{headergray}{RGB}{220,220,220}

\title{Improving Latent Reasoning in LLMs via Soft Concept Mixing}

\author{
Kang Wang$^{1}$, Xiangyu Duan$^{1}$\thanks{Corresponding author} , Tianyi Du$^{1}$ \\
\\
$^{1}$ School of Computer Science and Technology, Soochow University \\
\texttt{kwang0822@stu.suda.edu.cn}
}

\begin{document}

\maketitle
\footnotetext[1]{© 2026 IEEE. Personal use of this material is permitted. Permission from IEEE must be obtained for all other uses, in any current or future media, including reprinting/republishing this material for advertising or promotional purposes, creating new collective works, for resale or redistribution to servers or lists, or reuse of any copyrighted component of this work in other works.}

\begin{abstract}
Unlike human reasoning in abstract conceptual spaces, large language models (LLMs) typically reason by generating discrete tokens, which potentially limit their expressive power. The recent work Soft Thinking has shown that LLMs' latent reasoning via soft concepts is a promising direction, but LLMs are trained on discrete tokens. To reduce this gap between the soft concepts in reasoning and the discrete tokens in training, we propose Soft Concept Mixing (SCM), a soft concept aware training scheme that directly exposes the model to soft representations during training. Specifically, SCM constructs a soft concept vector by forming a probability-weighted average of embeddings. Then, this vector is mixed into the model's hidden states, which embody rich contextual information. Finally, the entire latent reasoning process is optimized with Reinforcement Learning (RL). Experiments on five reasoning benchmarks demonstrate that SCM improves the reasoning performance of LLMs, and simultaneously maintains a stable training dynamic.
\end{abstract}

% keywords can be removed

\section{Introduction}
\label{sec:intro}
Large language models (LLMs) excel in complex reasoning tasks like mathematics, commonsense, and code. Chain-of-Thought (CoT) prompting \cite{wei2022chain,lobo2024impact} and its related works \cite{guo2025deepseek,zhang2025self,zhang2022automatic} substantially improve LLM performance by decomposing problems step by step through the generation of intermediate reasoning traces. However, standard CoT reasoning constrains the reasoning process strictly to sequences of discrete tokens \cite{yao2023tree}. This limitation is not only constrained by the semantic granularity of natural language but also forces the model to advance along a single trajectory at each step, making it difficult to explore alternative plausible reasoning paths in parallel.

Unlike language models that generate conclusions word by word, human reasoning occurs in a high-dimensional abstract concept space before being expressed in language \cite{quiroga2005invariant,fedorenko2016language}. This process enables the simultaneous exploration of multiple possibilities, facilitating comparison of reasoning paths and reducing the risk of premature errors under uncertainty.

Recent studies have explored latent space reasoning \cite{zhu2025survey,xu2025softcot,cheng2024compressed,tack2025llm,yang2024large}. For instance, Coconut \cite{hao2024training} uses the last hidden state of the model as the next-step input embedding, thereby achieving a form of “continuous thought” within the latent space. However, its training paradigm relies on a complex, multi-stage training process that requires a large corpus of CoT trajectories for supervision. This approach is not only computationally intensive but also potentially degrades the general capabilities acquired during pre-training. In contrast, a Reinforcement Learning (RL) framework does not require CoT trajectories, allowing the model to freely explore the reasoning space and unlock its intrinsic potential.

Another line of work, Soft Thinking \cite{zhang2025soft} leverages the full output probability distribution to construct a concept token, allowing reasoning to proceed in a high-dimensional semantic space. However, as a method applied only at inference time, it introduces a fundamental gap between its continuous reasoning process and the model's training on discrete tokens. Since the model is never exposed to such continuous representations during its training phase, it struggles to utilize these soft concepts, causing performance instability. This requires the development of a lightweight framework that empowers the model to effectively explore and internalize continuous concepts, weaving them directly into the fabric of its own rich, contextual understanding during training.

To achieve this, we propose \textbf{S}oft \textbf{C}oncept \textbf{M}ixing (\textbf{SCM}), a novel framework that integrates soft concept vectors directly into the language model's training phase. At each reasoning step, SCM not only samples discrete tokens but also leverages the model’s full output probability distribution to generate a soft concept vector via probability-weighted average of embeddings. This vector is mixed into the model’s hidden states, preserving the distribution's rich latent semantics. To optimize the complex policy space that combines both discrete tokens and continuous concepts, we adopt a RL framework based on Group Relative Policy Optimization (GRPO) \cite{shao2024deepseekmath}, which stably optimizes the latent policy space and effectively guides the model’s internal reasoning process.

% Our evaluations on reasoning benchmarks show that SCM improves performance across different model scales. Principal Component Analysis (PCA) of hidden states further confirms the representational stability of our approach, showing only slight latent shifts, comparable to the GRPO baseline.
Our evaluations on reasoning benchmarks show that SCM improves performance across different model scales. Furthermore, our approach demonstrates high representational stability; Principal Component Analysis (PCA) of hidden states reveals only slight latent shifts, comparable to the GRPO baseline.
\begin{figure*}[tbp]
    \centering % 图片居中
    \includegraphics[width=1.0\textwidth]{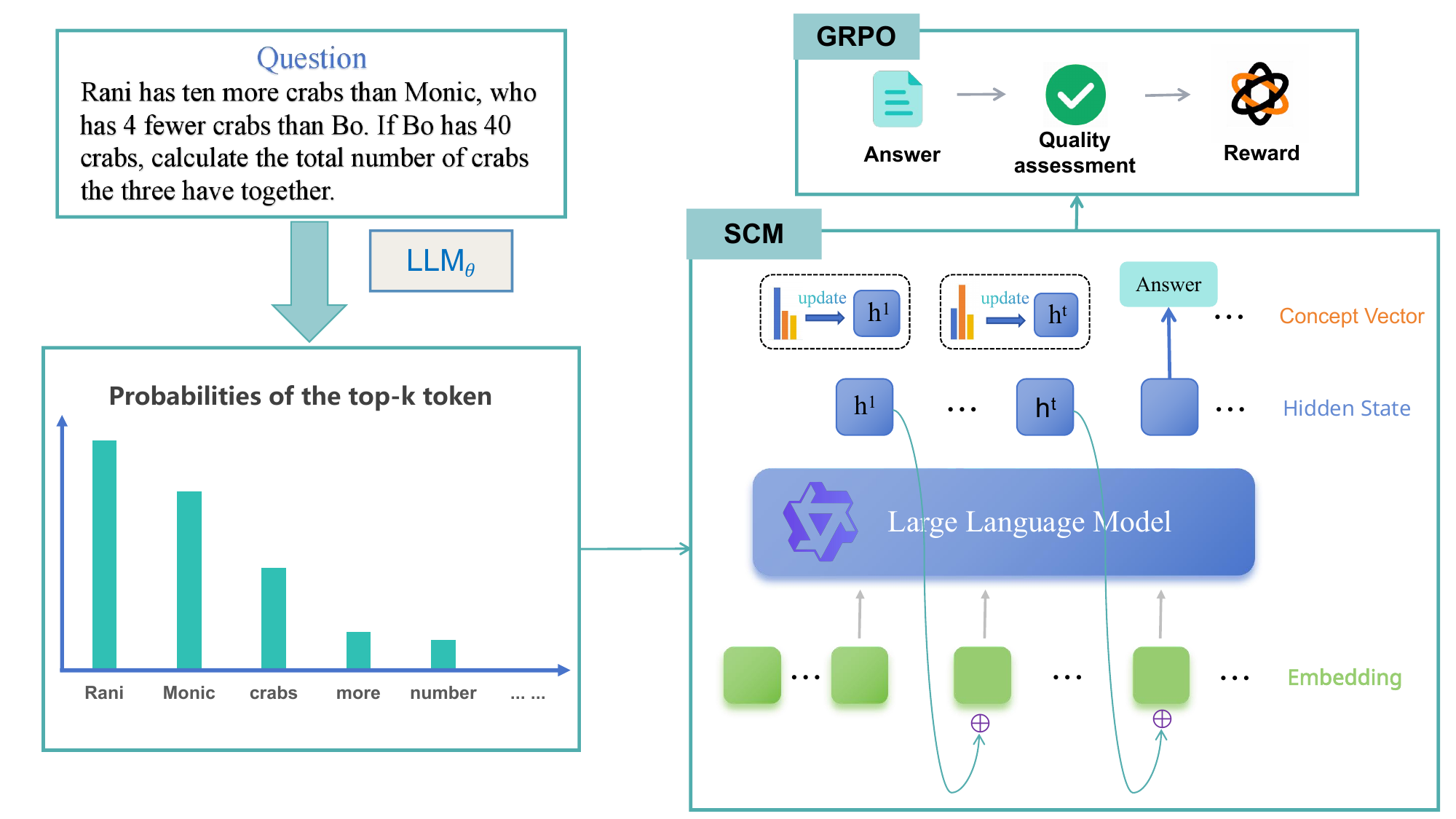} % 设置图片宽度为文本宽度（跨双栏宽度），指定图片路径
    \caption{Training Framework for Soft Concept Mixing (SCM).} % 图片标题
    \label{fig:framework} % 图片标签，用于文中引用
\end{figure*}

\section{PROPOSED METHOD}
\label{sec:PROPOSED METHOD}
In this section, we introduce Soft Concept Mixing (SCM), a novel training framework designed to enhance the latent reasoning capabilities of Large Language Models. The overall process is illustrated in Figure \ref{fig:framework}.

\subsection{Soft Concept Vector Generation}
To create a guidance signal that is more informative and fine-grained than any single discrete token, we introduce the soft concept vector.
At each decoding step \(t\), the model produces a probability distribution \(\boldsymbol{p}_t = \{p_{t,i}\}_{i=1}^{|V|}\) over its vocabulary \(V\). This distribution is then used to construct a probability-weighted average of embeddings \(E = \{\boldsymbol{e}(x_i)\}_{i=1}^{|V|}\), forming the soft concept vector \(\widetilde{\boldsymbol{se}}_t\):
\[
\widetilde{\boldsymbol{se}}_t = \sum_{i = 1}^{|V|} p_{t,i} \cdot \boldsymbol{e}(x_i).
\]

The resulting vector \(\widetilde{\boldsymbol{se}}_t\) serves as a compact, continuous representation of all latent reasoning paths at the current step.

\subsection{Integration with Hidden State}
\label{sec:integration}
Since the hidden state contains rich contextual information, we consider it the ideal integration point. In an autoregressive model, this hidden state \(\boldsymbol{h}_t\) is generated from the input \(\boldsymbol{x}\) and the preceding tokens \(\boldsymbol{y}_{<t}\) as follows:
\(
\boldsymbol{h}_t = \text{LLM}_\theta(\boldsymbol{y}_{<t}, \boldsymbol{x})
\)

We then integrate the soft concept vector \(\widetilde{\boldsymbol{se}}_t\) with the model's hidden state \(\boldsymbol{h}_t\) via simple addition: \(\boldsymbol{h}'_t = \boldsymbol{h}_t + \widetilde{\boldsymbol{se}}_t\). This parameter-free approach is chosen for its simplicity and efficiency, ensuring stability during RL optimization. Subsequently, the policy samples the next token conditioned on this enhanced hidden state:
\[
y_t \sim q_\theta(\cdot \mid x, y_{<t}, h'_t)
\]

\subsection{Optimization with Reinforcement Learning (GRPO)}
We train the above latent policy with GRPO~\cite{shao2024deepseekmath}. For each input $x$, we sample $K$ rollouts. In each of these rollouts, the model applies the concept-mixing operation at every step to obtain the mixed state \({h'_t}\) and then samples \(y_t \sim q_\theta(\cdot \mid x, y_{<t}, \{h'_t\})\).

We use a reward $r^{(k)}$ composed of a binary correctness term $r_{\text{acc}}^{(k)}$ and a structural format term $r_{\text{fmt}}^{(k)}$. Specifically, $r_{\text{acc}}$ is 1 for a correct final answer (0 otherwise), and $r_{\text{fmt}}$ grants +0.25 for each of the four required structural tags (\texttt{<think>}, \texttt{</think>}, \texttt{<answer>},\texttt{</answer>}). The total reward is defined as the sum of these components: $r^{(k)} = r_{\text{acc}}^{(k)} + r_{\text{fmt}}^{(k)}$. Instead of using an explicit value network, GRPO estimates the baseline using the group statistics. The advantage $A^{(k)}$ for the $k$-th output in a group of size $K$ is calculated as:
\[
A^{(k)} = \frac{r^{(k)} - \mathrm{mean}\left(r^{(1)}, r^{(2)}, \dots, r^{(K)}\right)}{\mathrm{std}\left(r^{(1)}, r^{(2)}, \dots, r^{(K)}\right) + \varepsilon},
\]
where $\varepsilon$ is a small constant for numerical stability.

Specifically, for each input query $x$, GRPO samples a group of $K$ outputs $\{y^{(1)}, y^{(2)}, \dots, y^{(K)}\}$ from the reference policy $\pi_{\theta_{\text{old}}}$. Following the formulation in Section~\ref{sec:integration}, we denote the generation probability of the $k$-th output sequence as $\pi_\theta(y^{(k)} \mid x)$. The policy $\pi_\theta$ is then optimized by minimizing the following clipped surrogate objective:

\[
\mathcal{L}(\theta) = -\frac{1}{K} \sum_{k=1}^{K} \left( \min \left( 
\frac{\pi_\theta(y^{(k)} \mid x)}{\pi_{\theta_{\text{old}}}(y^{(k)} \mid x)} A^{(k)}, \;
\text{clip} \left( \frac{\pi_\theta(y^{(k)} \mid x)}{\pi_{\theta_{\text{old}}}(y^{(k)} \mid x)}, 1-\varepsilon, 1+\varepsilon \right) A^{(k)} 
\right) \right),
\]

where $\varepsilon$ is a hyperparameter controlling the clipping range, and the ratio $\frac{\pi_\theta}{\pi_{\theta_{\text{old}}}}$ represents the importance sampling weight, which measures the deviation from the reference policy.

This objective yields stable gradients without a learned value function or KL regularization, and directly reinforces trajectories that achieve higher rewards under the concept-mixed policy.

\section{EXPERIMENTS}
\label{sec:experiment}
\subsection{Experimental setup}
\textbf{RL Setup}.
We fine-tune our models using the GRPO algorithm, implemented with the Unsloth framework. We apply a trainable LoRA module \cite{hu2022lora} (rank=32, alpha=64), targeting all linear layers in the attention (Q/K/V/O) and MLP (gate/up/down) blocks. The models are trained for one epoch on four NVIDIA A100 GPUs with a global batch size of 64, sampling $K=8$ rollouts per prompt. We use a learning rate of $5 \times 10^{-6}$ and a maximum output length of 4096 tokens. The system prompt
 adopted in the RL training is provided below.
% 创建一个 tcolorbox
\begin{tcolorbox}[
    colback=white,            % 内容背景色：白色
    colframe=gray!60,         % 边框颜色：深灰色
    colbacktitle=headergray,  % 标题栏背景色：浅灰色
    coltitle=black,           % 标题文字颜色：黑色
    fonttitle=\rmfamily,      % 标题字体：衬线体（类似Times）
    title={System prompt for RL training}, % 您的新标题
    arc=2mm,                  % 圆角弧度
    boxrule=0.5pt,            % 边框粗细
    left=3mm, right=3mm, top=3mm, bottom=3mm, % 内容边距
    toptitle=1.5mm, bottomtitle=1.5mm,        % 标题边距
    fontupper=\rmfamily       % 内容字体：衬线体
]
    % 这里是您的Prompt内容
    % 注意：LaTeX特殊符号如 \ { } 需要转义
    Think about the problem and provide your working out. Then put your final answer within \texttt{\textbackslash boxed\{\}}.
    The reasoning process and answer are enclosed within '\texttt{<think>}' '\texttt{</think>}' and '\texttt{<answer>}' '\texttt{</answer>}' tags, respectively, i.e., \texttt{<think>} \{reasoning process\} \texttt{</think>} \texttt{<answer>} \texttt{\textbackslash boxed\{\{final answer\}\}}  \texttt{</answer>}.
    
    \{Question\}
\end{tcolorbox}

\textbf{Dataset and Models}. We train on two mathematical reasoning datasets—GSM8K \cite{cobbe2021training} and MATH \cite{hendrycks2021measuring}, to cover a wide difficulty range from elementary arithmetic to competition-level problems. We conduct experiments using four open-source LLMs: DeepSeek-R1-Distill-Qwen-7B, DeepSeek-R1-Distill-Llama-8B, DeepSeek-R1-Distill-Qwen-1.5B \cite{guo2025deepseek}, and Qwen2.5-7B-Instruct \cite{qwen2}.

\textbf{Evaluation Settings}. We evaluate the models before and after reinforcement learning on five public reasoning benchmarks:
GSM8K~\cite{cobbe2021training}, MATH500~\cite{hendrycks2021measuring}, AIME~2024~\cite{li2024numinamath}, GPQA-Diamond~\cite{rein2024gpqa}, and MMLU~\cite{hendrycks2020measuring}.
All methods use the same chat templates and CoT prompting, and we report pass@1 accuracy. For decoding, we use top-$k{=}30$ sampling with temperature $0.6$, top-$p{=}0.95$, and a maximum generated length of $32{,}768$ tokens. Soft Thinking \cite{zhang2025soft} is evaluated with its official implementation and default hyperparameters. The system prompt provided in below.

% 创建一个 tcolorbox
\begin{tcolorbox}[
    colback=white,            % 内容背景色：白色
    colframe=gray!60,         % 边框颜色：深灰色
    colbacktitle=headergray,  % 标题栏背景色：浅灰色
    coltitle=black,           % 标题文字颜色：黑色
    fonttitle=\rmfamily,      % 标题字体：衬线体（类似Times）
    title={System prompt for evaluation on math  benchmarks}, % 您的新标题
    arc=2mm,                  % 圆角弧度
    boxrule=0.5pt,            % 边框粗细
    left=3mm, right=3mm, top=3mm, bottom=3mm, % 内容边距
    toptitle=1.5mm, bottomtitle=1.5mm,        % 标题边距
    fontupper=\rmfamily       % 内容字体：衬线体
]
    % 这里是您的Prompt内容
    % 注意：LaTeX特殊符号如 \ { } 需要转义
    Please reason step by step, and put your final answer within \texttt{\textbackslash boxed\{\{\}\}}.
    
    \{Question\}
\end{tcolorbox}

\begin{table*}[htbp]
  \centering
  \resizebox{1.0\textwidth}{!}{ % 缩放到合适宽度
  \begin{tabular}{llcccccc} % 列结构：Model(1列)+Method(1列)+5指标(5列)+Avg.(1列)，共8列
    \toprule[1.2pt]
    Model & Method & MATH 500 & AIME 2024 & GSM8k & GPQA-Diamond & MMLU & Avg. \\
    \midrule
    % 第一个模型：跨5行（4个方法+1个ablation），垂直居中
    \multirow{5}{*}{\centering DS-R1-Q-7B} 
      & CoT & 91.80 & 50.00 & 90.37 & 50.51 & 66.61 & 69.86 \\
      & Soft Thinking & 90.60 & 46.67 & 89.31 & 49.50 & 66.32 & 68.48 \\
      & GRPO & 93.20 & \textbf{56.67} & 90.52 & 51.01 & 66.83 & 71.65 \\
      % & HRPO & 93.8 & 56.67 & 89.54 & 50.51 & 66.66 & 71.44 \\
      
      & SCM & \textbf{94.40} & \textbf{56.67} & \textbf{92.03} & \textbf{51.52} & \textbf{66.98} & \textbf{72.32} \\
      & \quad w/o hidden states & 93.00 & \textbf{56.67} & 90.00 & 50.51 & 66.66 & 71.37 \\
    \midrule
    % 第二个模型：跨5行，垂直居中
    \multirow{5}{*}{\centering DS-R1-L-8B}
      & CoT & 81.20 & 43.30 & 69.29 & 44.44 & 71.09 & 61.86 \\
      & Soft Thinking & 77.20 & 40.00 & 70.05 & 42.93 & 70.50 & 60.14 \\
      & GRPO & 81.60 & 50.00 & 70.13 & \textbf{45.45} & \textbf{72.81} & 64.00 \\
      & SCM & \textbf{81.80} & \textbf{53.30} & \textbf{70.96} & 44.95 & 71.65 & \textbf{64.53} \\
      & \quad w/o hidden states & 81.60 & 50.00 & 70.43 & 44.85 & 71.32 & 63.64 \\
    \midrule
    % 第三个模型：跨5行，垂直居中
    \multirow{5}{*}{\centering DS-R1-Q-1.5B}
      & CoT & 80.80 & 30.00 & 75.21 & 33.33 & 44.38 & 52.74 \\
      & Soft Thinking & 80.60 & 20.00 & 77.33 & 31.82 & 44.17 & 50.78 \\
      & GRPO & \textbf{81.80} & 33.33 & 76.95 & 33.84 & 45.02 & 54.19 \\
      & SCM & 81.00 & \textbf{36.67} & \textbf{77.79} & \textbf{34.34} & 45.20 & \textbf{55.00} \\
      & \quad w/o hidden states & 80.60 & 33.33 & 77.18 & 33.84 & \textbf{45.80} & 54.15 \\
    \midrule
    % 第四个模型：跨5行，垂直居中
    \multirow{5}{*}{\centering Qwen2.5-7B-INS}
      & CoT & 74.80 & 13.33 & 89.61 & 32.32 & 68.29 & 55.67 \\
      & Soft Thinking & 75.80 & 13.33 & 89.76 & 32.32 & 68.19 & 55.88 \\
      & GRPO & 75.40 & 13.33 & 90.67 & \textbf{32.83} & 68.45 & 56.14 \\
      & SCM & \textbf{76.20} & \textbf{16.67} & \textbf{91.89} & \textbf{32.83} & \textbf{68.55} & \textbf{57.23} \\
      & \quad w/o hidden states & 75.80 & 13.33 & 90.37 & 32.32 & 68.09 & 55.98 \\
    \bottomrule[1.2pt]
  \end{tabular}}
  \caption{Accuracy of SCM compared to other baselines on five reasoning benchmarks. The GRPO baseline enhances CoT with RL fine-tuning, while SCM w/o hidden states is an ablation that removes hidden state fusion. Best results are in \textbf{bold}.}
  \label{tab:main_result}
\end{table*}

\subsection{Results and Analysis}
We evaluated the proposed Soft Concept Mixing (SCM) method on four models, and compared it with traditional Chain-of-Thought (CoT) reasoning, Soft Thinking, and the reinforcement learning baseline of GRPO fine-tuned on CoT. In addition, we conducted an ablation study by removing the hidden state fusion component (denoted as SCM w/o hidden states). Detailed results are shown in Table \ref{tab:main_result}.

% Our experiments show that Soft Concept Mixing (SCM) consistently outperforms strong baselines across different model scales and types, enhances training stability, and underscores the effectiveness of its core hidden state fusion mechanism. Detailed results are shown in Table \ref{tab:main_result}.

First and foremost, SCM demonstrates a significant performance advantage even over the strong GRPO reinforcement learning baseline, indicating that mixing soft concept vectors provides a more effective and stable guidance signal for the RL process. This improvement is also evident when compared to other methods; on challenging benchmarks like AIME 2024 and GSM8K, SCM significantly outperformed both CoT and Soft Thinking. The gain over Soft Thinking is particularly noteworthy, as it empirically confirms the benefit of resolving the train-inference mismatch by exposing the model to soft concept vectors during training.

\begin{figure}[htbp]
    \centering
    \includegraphics[width=0.6\linewidth]{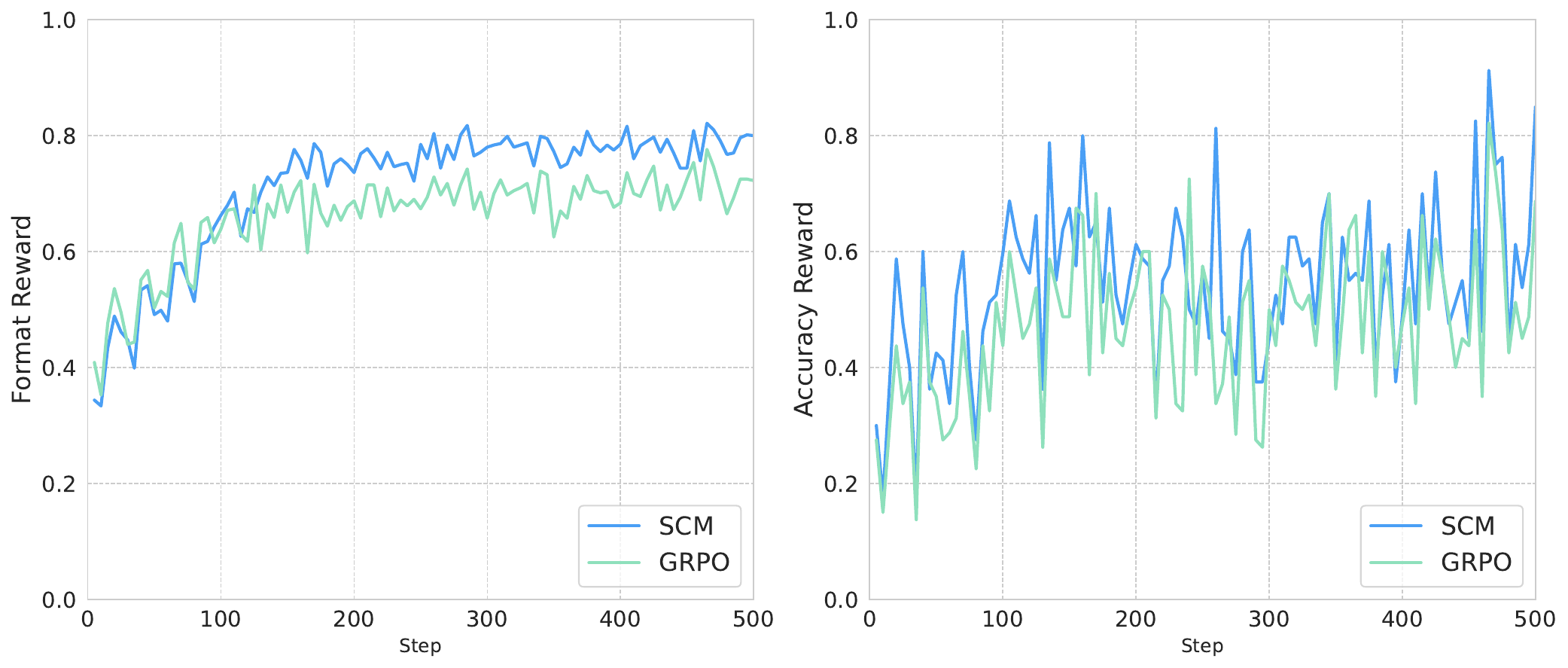} % 从0.8增加到0.95
    \caption{Training process of DeepSeek-R1-Distill-Qwen-1.5B on GRPO utilizing our proposed SCM method.}
    \label{fig:real_data_plot}
\end{figure}

Furthermore, the robustness of SCM is validated on models with limited capacity. On the lightweight DeepSeek-R1-Distill-Qwen-1.5B, SCM still surpasses CoT, Soft Thinking, and GRPO. This stability is also reflected in the training dynamics, where Figure \ref{fig:real_data_plot} shows SCM exhibits higher and more stable rewards compared to GRPO in later training stages. The method’s applicability extends beyond reasoning-oriented architectures, as SCM also yields consistent gains on a instruction-tuned model Qwen2.5-7B-Instruct. Finally, our ablation study confirms the necessity of the hidden state fusion component, as its removal degrades performance.

To compare with recent latent-reasoning approaches, we evaluate Coconut~\cite{hao2024training} and HRPO~\cite{yue2025hybrid}. To benchmark against Coconut, we adapted its official codebase from the original GPT-2 target to DeepSeek-R1-Distill-Qwen-7B. We then fine-tuned the model using a trainable LoRA module. For fairness, Coconut and HRPO are trained on the same corpora as SCM---GSM8K~\cite{cobbe2021training} and MATH~\cite{hendrycks2021measuring}, with HRPO using its official hyperparameters. At inference time, we keep the evaluation setup identical to our SCM runs (same chat template, decoding temperature, maximum output lengths, and pass@1). Results are reported in Table~\ref{tab:concurrent_methods}.

\begin{table}[htbp]
  \centering
  \begin{tabular}{lcccccc}
    \toprule
    Method       & MATH 500 & AIME 2024 & GSM8K & GPQA-Diamond & MMLU  & Avg   \\
    \midrule
    CoT(Baseline) & 91.80    & 50.00     & 90.37 & 50.51        & 66.61 & 69.44 \\
    HRPO         & 93.80    & \textbf{56.67}     & 89.54 & 50.51        & 66.66 & 71.86 \\
    Coconut      & 93.60    & 43.67     & 89.03 & 46.98        & 66.45 & 67.93 \\
    SCM          & \textbf{94.40}    & 53.33     & \textbf{92.31} & \textbf{51.52}        & \textbf{66.98} & \textbf{72.32} \\
    \bottomrule
  \end{tabular}
  \vspace{0.8em}
  \caption{Comparison with concurrent methods on DeepSeek-R1-Distill-Owen-7B}
  \label{tab:concurrent_methods}
\end{table}

\subsection{Latent Space Shift Analysis}
To further investigate the impact of SCM on the internal representation structure of models, we employ Principal Component Analysis (PCA) to visualize and quantitatively evaluate the shift of latent representation centers before and after training across different models \cite{huan2025does,zheng2025spurious,xu2025unlearning}. Specifically, we apply PCA ($n=2$) to the hidden states of all transformer layers for each model state (pre- vs. post-training), and compute the average PCA coordinate of each layer as the representation center $z^{(\cdot)}$. The latent space deviation between two model states is then defined as the Euclidean distance between their centers: $d^{(\cdot)} = \| z^{(\cdot)} - z^{(\text{orig})} \|_2$, where $z^{(\text{orig})}$ denotes the center of the initialized model.
\begin{table}[htbp]
  \centering
  \begin{tabular}{lccc}
    \toprule
    Model       & Method & Math Datasets & Commonsense Datasets \\
    \midrule
    DS-R1-L-8B  & GRPO   & 0.02          & 0.05                 \\
                & SCM    & 0.09          & 0.04                 \\
    \midrule
    DS-R1-Q-1.5B& GRPO   & 0.04          & 0.05                 \\
                & SCM    & 0.05          & 0.09                 \\
    \bottomrule
  \end{tabular}
  \vspace{0.8em}
  \caption{PCA shift distances of representation centers across models}
  \label{tab:pca_shift_metrics}
\end{table}

\begin{figure*}[htbp]  % t 表示优先放置在页面顶部（top），实现占双栏最上面
    \centering
    % width=0.8\textwidth 设为跨双栏宽度的0.8倍，默认保持纵横比
    \includegraphics[width=1.0\textwidth]{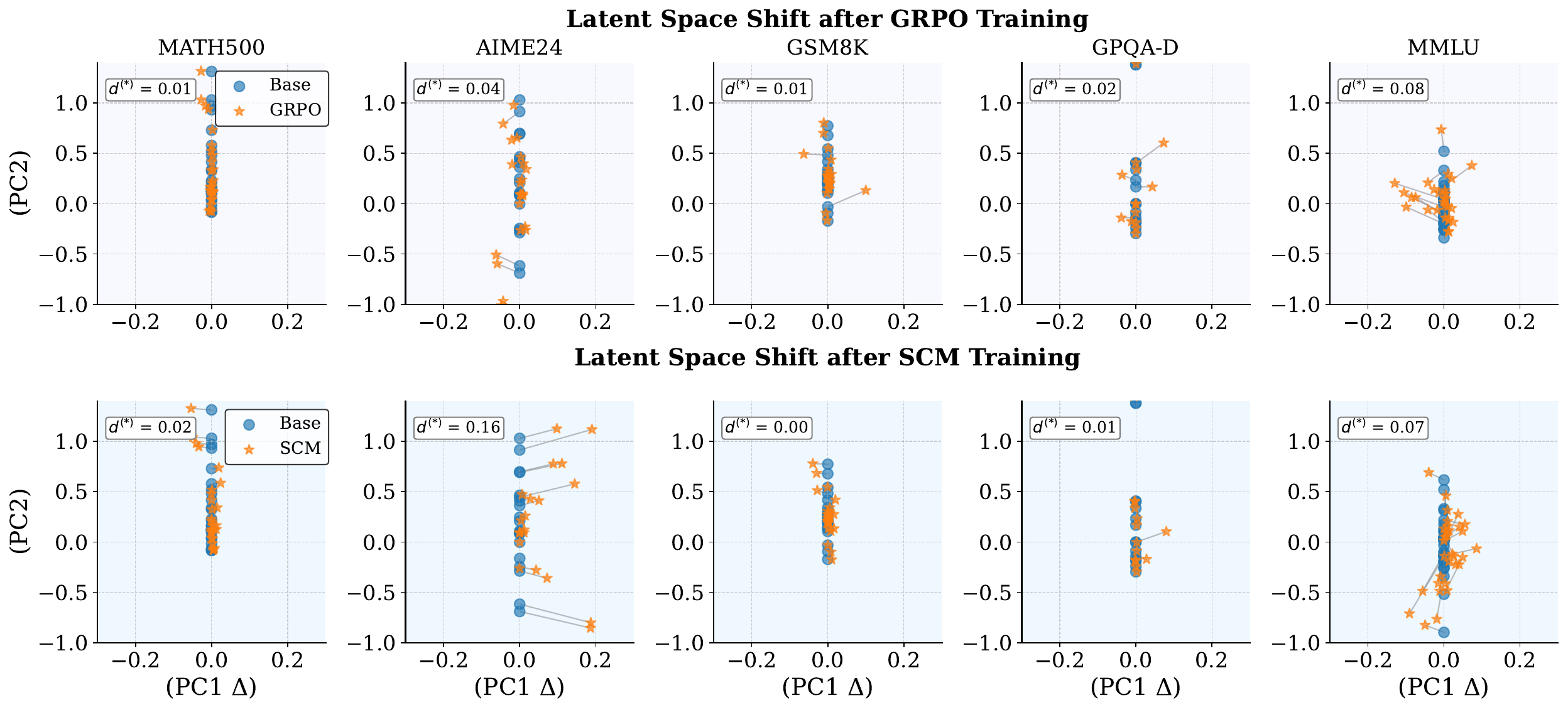} 
    \caption{PCA shift of DeepSeek-R1-Distill-Llama-8B across five datasets}
    \label{fig:pca_shift}
\end{figure*}

Figure \ref{fig:pca_shift} presents PCA distribution comparisons for the same model evaluated in five different datasets. We observe that whether trained solely with GRPO or with the additional SCM mechanism, the overall latent representation shift before and after training shows no significant large-scale drift, indicating stable policy optimization. This trend is consistent across different model architectures (Table \ref{tab:pca_shift_metrics}). On this basis, the SCM-enhanced models achieve higher accuracy across multiple tasks while maintaining a stable latent representation structure without significant drift after training, further demonstrating our method’s balanced trade-off between performance and representational stability.

\section{CONCLUSION}
\label{sec:conclusion}
In this work, we introduced Soft Concept Mixing (SCM), a training framework that resolves a key challenge in latent reasoning: the mismatch between models trained on discrete tokens and the need to reason in a continuous semantic space. We demonstrated that by constructing probability-weighted soft concept vectors and integrating them directly into the model's hidden states during reinforcement learning, SCM effectively bridges this train-inference gap. Our experiments across multiple reasoning benchmarks and model architectures have shown that SCM significantly enhances reasoning performance and maintains stable training dynamics.

\bibliographystyle{unsrt}  


\bibliography{references}

\end{document}